# Reinforcement Learning for Learning of Dynamical Systems in Uncertain Environment: A Tutorial

**Mehran Attar*[1] and Mohammad Reza Dabirian[2]**

[1] Department of Electrical and Computer Engineering,

Tarbiat Modares University (TMU), Tehran, Iran.

[2] Department of Electrical Engineering,

Hamedan University of Technology, Hamedan, Iran.

## 1  Abstract

In this paper, a review of model-free reinforcement learning for learning of dynamical systems in uncertain environments has discussed. For this purpose, the Markov Decision Process (MDP) will be reviewed. Furthermore, some learning algorithms such as Temporal Difference (TD) learning, Q-Learning, and Approximate Q-learning as model-free algorithms which constitute the main part of this article have been investigated, and benefits and drawbacks of each algorithm will be discussed. The discussed concepts in each section are explaining with details and examples.

## 2  Introduction

Reinforcement Learning (RL) is the training of machine learning models to make a sequence of decisions. The agent learns to achieve a goal in an uncertain, potentially complex environment. In reinforcement learning, artificial intelligence faces a game-like situation. The computer employs trial and error to come up with a solution to the problem [1]. To get the machine to do what the programmer wants, artificial intelligence gets either rewards or penalties for the actions it performs. Its goal is to maximize the total score. Although the designer sets the reward policy–that is, the rules of the game–he gives the model no hints or suggestions for how to solve the game. It's up to the model to figure out how to perform the task to maximize the reward, starting from totally random trials and finishing with sophisticated tactics and superhuman skills. By leveraging the power of search and many trials, reinforcement learning is currently the most effective way to hint the machine's creativity. In contrast to human beings, artificial intelligence can gather experience from thousands of parallel gameplays if a reinforcement learning algorithm is run on sufficiently robust computer infrastructure [2, 3]. Reinforcement Learning (RL) constitutes a significant aspect of the Artificial Intelligence field with numerous applications ranging from finance, mathematics, state estimation theory to robotics, autonomous cars and a plethora of proposed approaches [4 - 10]. Furthermore, RL can increase the adaptability of dynamical systems, which is an important feature in order to deal with a complex and dynamic environment. One of the primary goals of the field of artificial intelligence.

---

*[1] Corresponding Author: mehran.attar@modares.ac.ir, **webpage:** www.mehranattar.com

[2]  Dabirianmohamadrez@gmail.com

In this manuscript, the main concentration is the uncertain searching for real dynamical systems, for instance, it is expecting that a robot learning to find its path without crashing with other objects in an environment or an autonomous car learning to set its speed in various situations. In certain searching, it is considered that the environment does not contain any noise and hence, the result of each action will be clear. However, in real environment the result of each action could be a probabilistic affair. The result of each action can correspond to a probabilistic function which calls transition function or dynamic of the system. If the dynamic of the system be a certain function, the problem will be a model-based, and if the dynamic of the system is an unknown function, the problem will be a model-free RL. The formulation of the problem in RL performs with Markov Decision Process (MDP). In other words, the agent - which can be a robot or an autonomous vehicle can perceive the problem of uncertain searching using MDPs. The interaction environment for the agent includes states and each state has a value. By taking various actions, the agent can change its' state. Figure 1 shows a grid environment for an agent. According to the Figure 1, there is 12 states, and agent start its's search at (1,1) state. Agent will choose the up action for 0.8 probability will go up and by choosing left and right action for 0.1 probability will go to left and right. The existence of probability could due to the noise in environment. By doing each action and going to the next state, the agent gets a feedback from the environment and this feedback is called reward which can be a positive or negative score. Besides, there are step rewards and overall reward. It is necessary to model the problem for the agent and as it is mentioned, this affair performs by MDP. In the next section, the foundation of MDP has introduced.

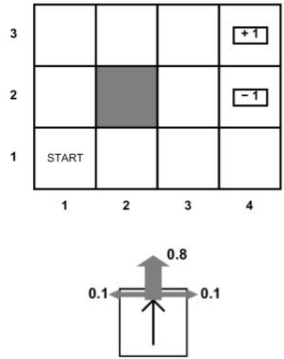

Figure 1: Grid Environment for an agent

## 3  Markov Decision Process

The elements of the RL can be formulized using the Markov Decision Process (MDP) framework. This section describes components such as states, actions, rewards and policies, as well as the goals of learning using different kinds of optimality criteria. MDPs are extensively described in [11, 12]. MDPs consist of states, actions, transitions between states and a reward function definition [13].

### 3.1  State

The set of environmental states $S$ is defined as the finite set $\{s^1, \dots, s^N\}$. where the *size* of the state space is $N$, i.e. $|S| = N$. A state is a unique characterization of all that is important in a state of the problem that is modelled. Some states may unachievable, e.g., in Figure 1 state (2,2) is unachievable.

### 3.2  Action

The set of actions $A$ is defined as the finite set $\{a^1, \dots, a^K\}$ where the *size* of the action space is $K$, i.e. $|A| = K$. Actions can be used to *control* the system state. The set of actions that can be applied in some particular state, $s \in S$, is denoted $A(s)$, where $A(s) \subseteq A$. In some systems, not all actions can be applied in every state, but in general we will assume that $A(s) = A$ for all $s \in S$.

### 3.3 Transition function

By applying action $a \in A$ in a state, $s \in S$, the system makes a transition from $s$ to a new state $s' \in S$, based on a probability distribution over the set of possible transitions. The transition function $T$ is defined as $T: S \times A \times S \rightarrow [0,1]$, i.e. the probability of ending up in state $s'$ after doing action, $a$ in state, $s$ is denoted $T(s, a, s')$. It is required that for all actions $a$, and all states $s$ and $s'$, $T(s, a, s') \geq 0$ and $T(s, a, s') \leq 1$. Furthermore, for all states $s$ and actions $a$, $\sum_{s' \in S} T(s, a, s') = 1$, $T$ defines a proper probability distribution over possible next states. Instead of a precondition function, it is also possible to set $T(s, a, s') = 0$ for all states $s' \in S$. if $a$ is not applicable in $s$. For talking about the *order* in which actions occur, we will define a discrete global clock, $t = 1, 2, \ldots$. Using this, the notation $s_t$ denotes the state at time $t$ and $s_{t+1}$ denotes the state at time $t + 1$. This enables to compare different states (and actions) occurring ordered in time during interaction. In some references, the transition functions call dynamic of the system.

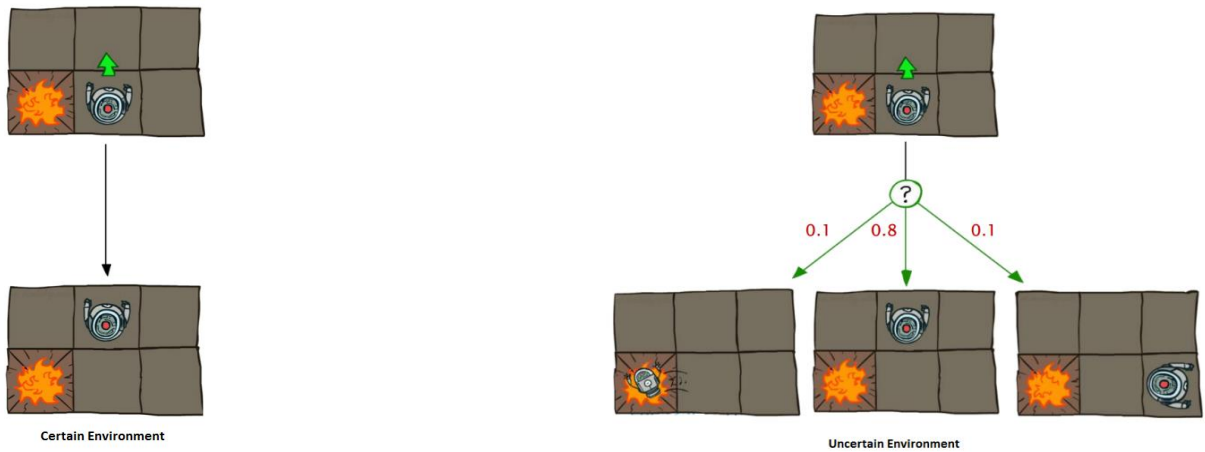

**Figure 2: Concept of Transition function-Agent in a certain and uncertain environment**

The system being controlled is Markovian if the result of an action does not depend on the previous actions and visited states (history), but only depends on the current state, i.e.

$P(s_{t+1}|s_t, a_t, s_{t-1}, a_{t-1}, \ldots) = P(s_{t+1}|s_t, a_t) = T(s_t, a_t, s_{t+1})$

The idea of Markovian dynamics is that the current state, s gives enough information to make an optimal decision; it is not important which states and actions preceded s. Another way of saying this, is that if you select an action a, the probability distribution over next states is the same as the last time you tried this action in the same state. More general models can be characterized by being k-Markov, i.e. the last k states are sufficient, such that Markov is actually 1-Markov. Though, each k-Markov problem can be transformed into an equivalent Markov problem.

### 3.4 Reward Function

The reward function specifies rewards for being in a state, or doing some action in a state. The state reward function is defined as $R: S \rightarrow R$, and it specifies the reward obtained in states. However, two other definitions exist. One can define either $R: S \times A \rightarrow R$ or $R: S \times A \times S \rightarrow R$. The first one gives rewards for performing an action in a state, and the second gives rewards for particular transitions between states. Throughout this book we will mainly use $R(s, a, s')$. The reward function is an important part of the MDP that specifies implicitly the *goal* of learning. For example, in episodic tasks such as in the games Tic-Tac-Toe and chess, one can assign all states in which the agent has won a positive reward value, all states in which the agent loses a negative reward value and a zero reward value in all states where the final outcome of the game is a draw. The goal of the agent is to reach positive valued states, which means winning the game. Thus, the reward function is used to give direction in which way the system, i.e. the MDP, should be controlled. Often, the reward function assigns non-zero reward to non-goal states as well, which can be interpreted as defining sub-goals for learning [14].

**Definition1:** A Markov decision process is a tuple S, A, T, R in which S is a finite set of states, A, finite set of actions, T a transition function defined as T: S×A×S → [0,1] and R a reward function defined as R: S×A×S → R. The transition function T and the reward function R together define the model of the MDP. Often MDPs are depicted as a state transition graph where the nodes correspond to states and (directed) edges denote transitions. A typical domain that is frequently used in the MDP literature is the maze Matthews (1922), in which the reward function assigns a positive reward for reaching the exit state [15].

### 3.5 Policy

Given an MDP $S, A, T, R$, a policy is a computable function that outputs for each state, $s \in S$ an action $a \in A$ (or $a \in A(s)$). Formally, a deterministic policy $\pi$ is a function defined as $\pi : S \rightarrow A$. It is also possible to define a stochastic policy as $\pi : S \times A \rightarrow [0,1]$ such that for each state, $s \in S$, it holds that $\pi(s, a) \geq 0$ and $\sum_{a \in A} \pi(s, a) = 1$.

Application of a policy to an MDP is done in the following way. First, a start state $s_0$ from the initial state distribution $I$ is generated. Then, the policy $\pi$ suggest the action $a_0 = \pi(s_0)$ and this action is performed. Based on the transition function $T$ and reward function $R$, a transition is made to state $s_1$, with probability $T(s_0, a_0, s_1)$ and a reward $r_0 = R(s_0, a_0, s_1)$ is received. This process continues, producing $s_0, a_0, r_0, s_1, a_1, r_1, s_2, a_2, ....$ If the task is episodic, the process ends in state $s_{goal}$ and is restarted in a new state drawn from $I$. If the task is continuing, the sequence of states can be extended indefinitely. The policy is part of the agent and its aim is to *control* the environment modelled as an MDP. A fixed policy induces a stationary transition distribution over the MDP which can be transformed into a Markov system $\langle S', T' \rangle$ where $S' = S$ and $T'(s, s') = T(s, a, s')$ whenever $\pi(s) = a$ [16].

### 3.6 Optimality Criteria and Discounting

In certain environments, we are looking for a chain of actions, however, in uncertain environments we should find a policy. As it is mentioned in previous section, policy is a function from states to actions. A policy is optimal if it gives us maximum benefit. An optimal policy defines as follows [17]:

$\pi^* : S \rightarrow A$

To make it obvious, an example of optimal policy has been provided. Consider an uncertain environment which has shown in Figure 3-a.

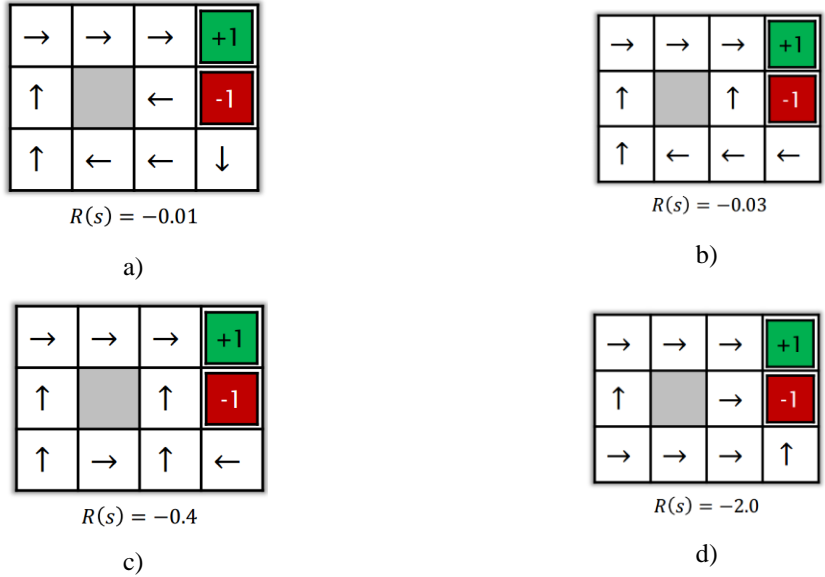

Figure 3: Concept of optimal policy

Each action has a reward of $R(s) = -0.01$. The best action in each state has shown. In Figure 3-b each action has a reward of $R(s) = -0.03$, and as a result, the policy will change. This change is due to the transition function, and therefore, the optimal policy is related to the transition function and reward function.

### 3.7 Concept of Q-State

If the agent is in the state s and choose the action a and perform the action a, it will go to state $s'$. On the other hand, if the agent just choose the action a, it will go to an imaginary state which call Q-state and it has shown as a green circle in Figure 4. This concept could be beneficial, especially in Temporal Difference (TD) learning and Q-Learning [18].

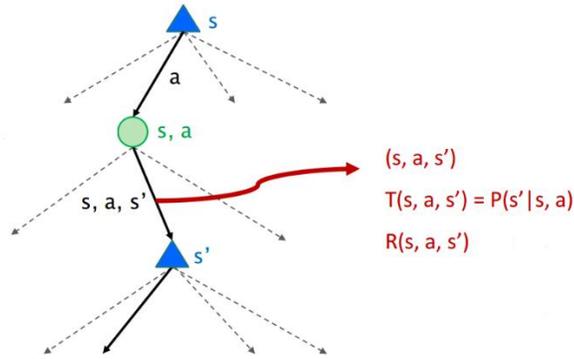

Figure 4: Concept of Q-State

MPDs are tools for modeling uncertain searching problems for an agent. One of the algorithms for solving a MDP is the Expectimax algorithm [19]. Even though this algorithm can solve low dimension problems, but it is not applicable for real problems because of high computational burden due to the infinite environment. For instance, in Figure 1, we have 11 states and 4 actions in each state. In real problems the number of state and actions can be an infinite number. To solve MDPs with infinite states and actions some algorithms have been proposed such as Approximate Q-Learning, which has investigated in this manuscript [20].

### 3.8 Value Functions and Bellman Equations

In the preceding sections we have defined MDPs and optimality criteria that can be useful for learning optimal policies. In this section, value functions will be defined, which are a way to link the optimality criteria to policies. Most learning algorithms for MDPs compute optimal policies by learning value functions. A value function represents an estimate how good it is for the agent to be in a certain state (or how good it is to perform a certain action in that state). The notion of how good is expressed in terms of an optimality criterion, i.e. in terms of the expected return. Value functions are defined for particular policies [21].

The value of a state, $s$ under policy $\pi$, denoted $V^{\pi}(s)$ is the expected return when starting in $s$ and following $\pi$ thereafter. We will use the infinite-horizon, discounted model in this section, such that this can be expressed as:

$$V^\pi(s) = E_\pi\{\sum_{k=0}^{\infty}(\gamma^k r_{t+k}|s_t = s)\} \qquad (3)$$

A similar state-action value function $Q: S \times A \rightarrow R$ can be defined as the expected return starting from state, $s$, taking action $a$ and thereafter following policy $\pi$[3]:

$$Q^\pi(s,a) = E_\pi\{\sum_{k=0}^{\infty}(\gamma^k r_{t+k}|s_t = s, a_t = a)\} \qquad (4)$$

One fundamental property of value functions is that they satisfy certain recursive properties. For any policy $\pi$ and any state, $s$ the expression in Equation (3) can recursively be defined in terms of a so-called Bellman Equation Bellman (1957):

$$V^\pi(s) = E_\pi\{r_t + \gamma r_{t+1} + \gamma^2 r_{t+2} + \cdots |s_t = t\} = E_\pi\{r_t + \gamma V^\pi(s_{t+1})|s_t = s\}$$
$$= \sum_{s'} T(s, \pi(s), s')(R(s, a, s') + \gamma V^\pi(s')) \qquad (5)$$

It denotes that the expected value of state is defined in terms of the immediate reward and values of possible next states weighted by their transition probabilities, and additionally a discount factor. $V^\pi$ is the unique solution for this set of equations.

Note that multiple policies can have the same value function, but for a given policy $\pi$, $V^\pi$ is unique. The goal for any given MDP is to find a *best* policy, i.e. the policy that receives the most reward. This means maximizing the value function of Equation (6), for all states $s \in S$. An optimal policy, denoted $\pi^*$, is such that $V^{\pi^*}(s) \geq V^\pi(s)$ for all $s \in S$ and all policies $\pi$. It can be proven that the optimal solution $V^* = V^{\pi^*}$ satisfies the following Equation:

$$V^*(s) = \max_{a \in A} \sum_{s' \in S} T(s, a, s')(R(s, a, s') + \gamma V^*(s')) \qquad (6)$$

This expression is called the Bellman optimality equation. It states that the value of a state under an optimal policy must be equal to the expected return for the best action in that state.

---

[3] $E_\pi$ shows the expected value under policy $\pi$.

### 3.9 Online and Offline MDP

In a MDP problem if the transition function and the policy are definite, the problem will be an offline MDP. In an online MDP, the transition and policy functions are not definite and the agent should learn them by interacting in the environment. In fact, by performing various episodes the agent tries to learn these functions.

### 3.10 Model-Base and Model-Free Learning

In model-based learning, the transition function is an unknown function and the agent by gaining experiences tries to estimate the transition function. If $T(s, a, s')$ is the transition function of an specific environment, the estimated transition function will be shown as $\hat{T}(s, a, s')$. The reward function is estimating as well. The estimated reward function is showing as $\hat{R}(s, a, s')$. To make it obvious, an example has been provided. Figure 5 shows a grid environment with predefined policy. The discount factor, $\gamma = 1$. By assuming that the agent has experienced four episodes, the estimated transition and reward function could be achieved according to these episodes.

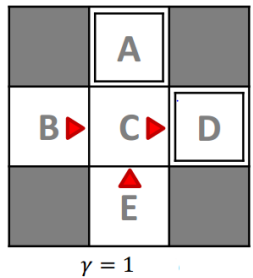

**Figure 5: Model-based Learning**

The episodes are as follows:

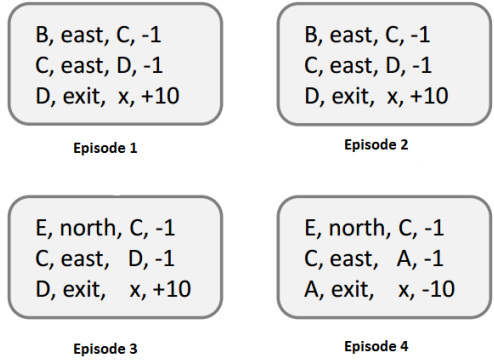

These four episodes have achieved by the agent, and the agent estimates transition and reward according to them. For instance, by starting from state B and taking east action, in the both episode 1 and 2 the agent has gone to the state C and has gotten a reward with score -1. Therefore, $\hat{T}(B, east, C) = 1$ and $\hat{R}(B, east, C) = -1$. By starting from state C and taking east action, in three episodes from four episodes, the agent has gone to the state D and has gotten a reward with score -1, therefore, $\hat{T}(C, east, D) = 0.75$ and $\hat{R}(C, east, D) = -1$. Finally, $\hat{T}(C, east, A) = 0.25$.

In model-free learning methods, the transition function has not estimated and instead the value of each state is estimated. In this manuscript, our concentration is on the model-free algorithms for solving MDP problems. The model-free algorithms are classifying in active and passive methods which has investigated in the next sub section.

### 3.11 Active and Passive Learning

In passive learning methods, the actions are imposing on the agent. In other words, the policy has predefined, and the agent acts according to this policy. As a result, the policy function, $\pi(s)$ is known. In passive learning method, the main goal of the agent is learning the value of each state. Direct evaluation is one of the passive methods which calculates the value of each state. To make it clear, an example has provided. Consider a grid environment which has shown in Figure 5. As the method of learning is a passive method, therefore, the policy of each state has determined in them by red arrows. By performing some episodes, the agent gains experiences and according to them, the agent learns the value of each state. The sample episodes are as follows:

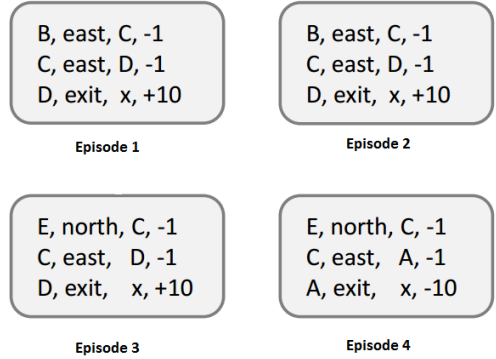

Episode 1: B, east, C, -1 / C, east, D, -1 / D, exit, x, +10

Episode 2: B, east, C, -1 / C, east, D, -1 / D, exit, x, +10

Episode 3: E, north, C, -1 / C, east, D, -1 / D, exit, x, +10

Episode 4: E, north, C, -1 / C, east, A, -1 / A, exit, x, -10

The value of each state is the summation of rewards from the that state to the final state. For instance, the value of state D is +10 because state D is the final state and in episodes 1, 2 and 3 by starting from D, the agent has gotten a reward with score of +10. The value of state A is -10 because state A is the final state and in episode 4 by starting from A, the agent has gotten a reward with the score of -10. It is necessary to mention that in this example the value of discount factor has considered $\gamma = 1$, to have more simple calculation. For the C state, there are four samples in episodes 1 to 4, and the value of state C will calculate according to these samples. The value of state C according to episodes 1, 2, 3 is +9 and according to episode 4 is -11. Therefore, final value of C will be +4. The value of other states is calculated respectively and has shown in Figure 6. Direct evaluation is a model-free algorithm because the value of each state is calculating without the calculation of transition and reward function. If the number of episode are enough, the value of states will converge to their actual value. Direct evaluation has some disadvantageous, for instance, the relation between states are not consider, it means that the learning of each state is an independent affair, therefore, the level of computational burden will rise and the agent need more time for learning each state.

**Figure 6: Direct evaluation method for calculating the value of each state**

One of the solutions for finding the relation between each state could be using policy evaluation method. According to the Bellman equation, the value of each state is calculated as follows:

$$V_0^\pi(s) = 0$$

$$V_{k+1}^\pi(s) \leftarrow \sum_{s'} T(s, a, s')\big(R(s, a, s') + \gamma V_k^\pi(s')\big) \qquad (7)$$

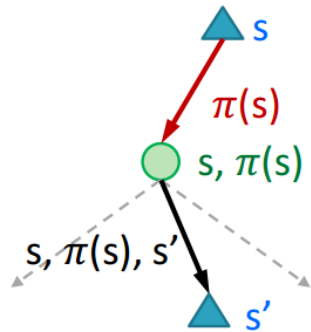

**Figure 7: Policy evaluation method**

According to (7), the relation between states for a fix policy has considered in policy evaluation method, but the main drawback of this method is that we should know transition and reward functions, and we want to solve MDP in an unknown environment without knowing these functions. Therefore, the key question is how the value of each state can calculate without having information about the transition and reward function. For solving this question some algorithms have been proposed, and the most important of them are the Temporal Difference (TD) learning and Q-learning. These algorithms will be discussed in the following sections.

## 4   Temporal Difference (TD) learning

The idea of TD learning is that the value of each state can estimate and update according to each sample that the agent achieving during the exploration in the environment. This process will continue until the estimated value of each state converge to the actual value of that state. By and large, the main idea of TD algorithm is learning based on each samples. In TD algorithm by getting each sample, the transition function is considering autonomously. According to Figure 8, if the agent is in the state $s$, and take the policy $\pi(s)$, then it will go to state $s'$ and gets the reward $R(s, \pi(s), s')$, and this transition is considered as a sample and is as follows:

$$sample = R(s, \pi(s), s') + \gamma V^\pi(s') \tag{8}$$

It should mention that transition from $s$ to $s'$ is probabilistic phenomenon. The achieved sample has observed and by using an interpolation, the convergence of estimated value to the real will achieve. This interpolation is as follows:

$$V^\pi(s) \leftarrow (1 - \alpha)V^\pi(s) + (\alpha)sample \tag{9}$$

In (9), $\alpha$ is the learning rate, and setting larger $\alpha$ means that the algorithm is more care about the achieved samples, and lower $\alpha$ means that the algorithm is more care about the estimated value. By rewriting (9), the following equation will achieve:

$$V^\pi(s) \leftarrow V^\pi(s) + (\alpha)(sample - V^\pi(s)) \tag{10}$$

In (10), the term $e = sample - V^\pi(s)$, is the estimation error of observation, and if $\lim_{k \to \infty} e = 0$ has converged.

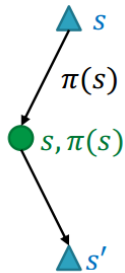

**Figure 8: Concept of TD algorithm**

**Remark 1** The role of updating values by using interpolation is calculating according to (11).

$$\bar{x}_n = (1 - \alpha).\bar{x}_{n-1} + \alpha.x_n \tag{11}$$

In (11), $\bar{x}_n$ is the current estimation, and $\bar{x}_{n-1}$ is the previous estimation. By rewriting (11), the following equation will achieve;

$$\bar{x}_n = \frac{x_n + (1-\alpha).x_{n-1} + (1-\alpha)^2.x_{n-2} + \cdots}{1 + (1-\alpha) + (1-\alpha)^2 + \cdots} \tag{12}$$

According to (12), if learning rate, $0 \leq \alpha \leq 1$ is more closer to 1, current samples are considered with more importance. In other words, the algorithm is going to forget past samples. In order to make the concepts of TD algorithm more clear, an example has been provided. Figure 9, shows an environment which the agent wants to explore in it and find the values of each state. To make the computations more simple, $\gamma = 1, \alpha = \frac{1}{2}$. There are 5 states at the beginning (A, B, C, D, E), and state, A and state, D are the final states. The initial values of each state is considered 0. By performing an action in each state, the agent will update the value of that state. The updating algorithm is as follows:

$$V^\pi(s) \leftarrow (1-\alpha)V^\pi(s) + (\alpha)[R(s,\pi(s),s') + \gamma V^\pi(s')] \tag{13}$$

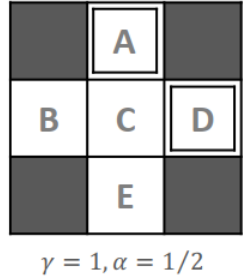

$\gamma = 1, \alpha = 1/2$

**Figure 9: TD algorithm concept-example.**

As it is mentioned before, the value of each state considered zero and the value of final states have determined in advanced. Therefore, in this example the value of states A to E have shown in Figure 10. It should be noted that if the agent is in state $s$ and perform action $a$, and go to state $s'$, then the value of state $s$ will update. This affair is also clear, by considering (13).

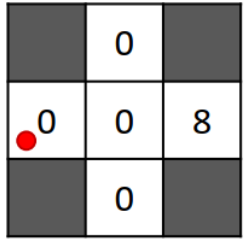

Figure 10: TD algorithm example.

According to Figure 10, the agent is in state B, and performing action (B, east, C, -2). By using (13), the new value of state B will be -1. This operation has shown in Figure 11. The agent is now in state C and performing action (C, east, D, -2), and the new value of state C will be 3. The operation has shown in Figure 12

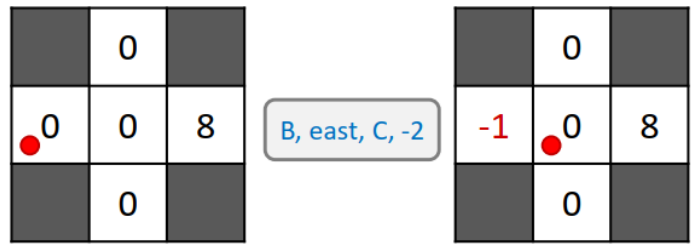

Figure 11: TD algorithm example.

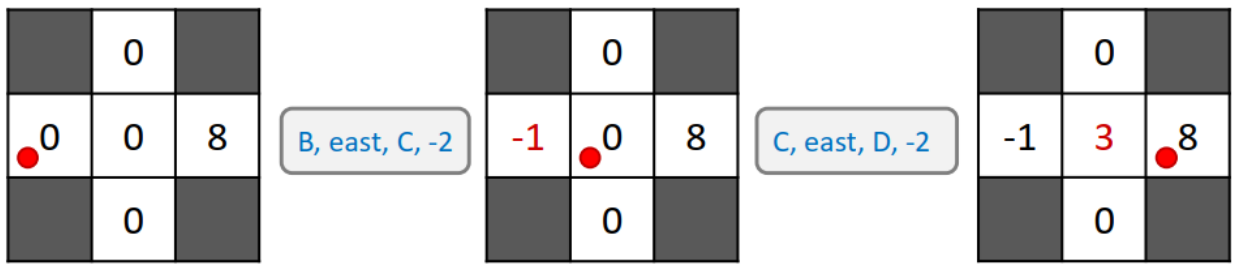

Figure 12: TD algorithm example.

Finally, by calculating the value of each state, the policy will achieve. TD algorithm is a model-free algorithm for policy evaluation. After calculation of each value state, the agent should choose an action, but for choosing an action, the agent needs to have information about the dynamic of the system, because the policy is achieving as follows:

$$Q(s,a) = \sum_{s'} T(s,a,s')[R(s,a,s') + \gamma V(s')]$$

$$\pi(s) = \arg\max_{a} Q(s,a)$$

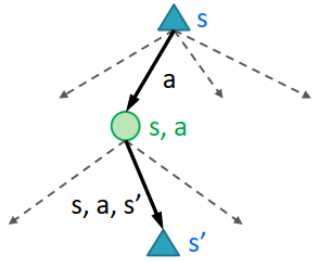

**Figure 13: Transition from state, *s* to *s'***

Without having the dynamic of the system, finding the optimal policy will be impossible, and this affair is considered as one of the drawbacks of TD algorithm. In fact, TD algorithm, is a model-free algorithm for finding the value of each state. The solution for solving this problem will be the calculation of optimal policy using the value of Q-states instead of states. In fact, each state is divided to Q-states, and in each Q-state includes the possible actions. Therefore, determining the optimal policy would be simpler by using Q-states, and the agent updates the value of Q-states by performing each action. This algorithm is called Q-learning and in the following this algorithm has been investigated.

## 5    Q-Learning Algorithm

In this section, the concepts of Q-learning algorithm for finding the optimal policy has presented. By using Q-learning, finding the suitable action will be performed without using the dynamic of the system. Q-learning is an active learning method which the agent has the freedom of taking action is each state. Accordingly, the following assumptions is considering in Q-learning algorithm:

- The dynamic of the system or the transition function $T(s, a, s')$ is unknown.
- The reward function $R(s, a, s')$ is unknown.
- The agent chooses each action autonomously.

As it is mentioned before, the main purpose in Q-learning algorithm is finding the optimal policy, however, adjustment between exploration and exploitation is a significant affair in this learning algorithm as well.

The initial values for each Q-state is set as $Q_0(s, a) = 0$, and in each iteration by having the vector $Q_k(s, a)$, the $Q_{k+1}(s, a)$ will calculate. In Q-learning algorithm, by receiving a sample $(s, a, s', r)$, and considering previous sample $Q(s, a)$, the new sample is calculating as follows:

$$sample = R(s, a, s') + \gamma \max_{a'} Q(s', a') \tag{15}$$

The equation (15), shows the estimation of new sample. In the next step, by using an interpolation between this new sample and previous sample, the estimated value of each Q-state converges to its real value. Therefore, the updating equation of estimation is as follows:

$$Q(s, a) = (1 - \alpha)Q(s, a) + (\alpha)[sample] \tag{16}$$

To make it clear, consider the following example. There is grid environment which has shown in Figure 14. In each state 4 actions are possible for the agent[4]. Hence, in each state there are four possible Q-states. The value of each Q-state will perform using Q-learning algorithm. Finally, in each state, those Q-state which has higher values are considered for the optimal action and will use to provide the optimal policy. In this example (Figure 14), the value of each Q-state has achieved after 1000 episodes of learning process.

---

[4] Actions are: north, left, right, south.

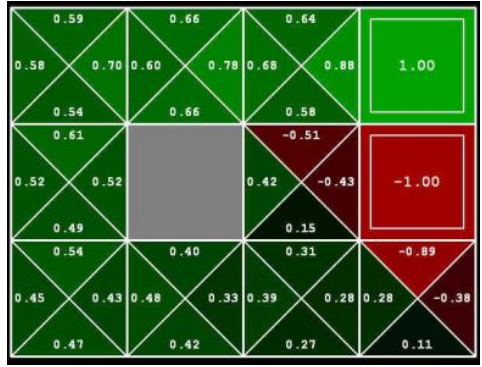

**Figure 14: grid environment- example for Q-learning**

**Remark 2** at the beginning of the learning, the value of learning rate should be high value and finally, it should decrease during the process of learning.

One of the main drawbacks of Q-learning algorithm is that this algorithm could be beneficial for finite number of states and actions, however, in real problems such as autonomous cars or robots, the number of states and actions are infinite and as a result, the computational burden of the Q-learning algorithm will increase dramatically. In this kind of problems, another algorithm has been used which calls Approximate Q-learning. In the following section, this algorithm has presented.

### 5.1   Importance of Exploration and Exploitation

One of important aspects in learning algorithms is the sufficient amount of exploration and exploitation which the agent should perform. If the agent does not explore enough in the environment, the probability of finding a global optimum will be very low and most of the time a local optimum will be found. Consider the grid environment which has shown in Figure 15. This environment has 17 states and 4 actions in each state. The start point is (1,2) state which has the reward of +1. There are 12 final states in top and blow of the grid environment which have -10 reward and 1 final state in (7,2) which has +10 reward. According to Figure 15, if the agent is in state (2,2), by performing low exploration, in most of the will find state (2,1) as an optimum solution and does not try other state. Therefore, the global optimum (state (7,2)) will never detect by the agent. As a result, the sufficient amount of exploration in the environment can give rise to finding global optimum. In order to have a sufficient exploration, the agent can use $\varepsilon - greedy$ exploration method which has presented in next section. Another method of having a sufficient exploration is the using of exploration functions. The main idea in exploration functions is that if a state has a bad score of reward, do not explore these states again and instead explore the new states. Exploration functions have presented in section 5.3. In the beginning of learning process, the value of $\varepsilon$ is set to low value and after performing enough episodes the value of $\varepsilon$ should set to zero, because the agent has learned the real value of each Q-state, and besides it can find the global optimum.

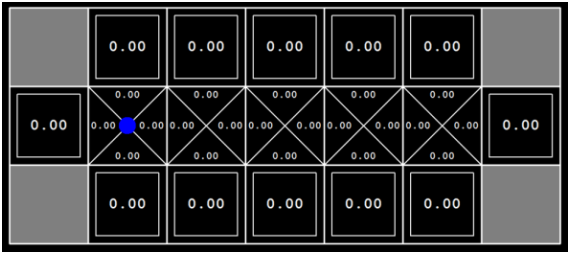

**Figure 15: grid environment- importance of exploration and exploitation**

### 5.2  $\varepsilon - greedy$ Exploration

To have a sufficient exploration, the agent can follow an $\varepsilon - greedy$ exploration method. In normal condition, the agent chooses its action according to the max value of each Q-state, however, in $\varepsilon - greedy$ exploration, the agent chooses its action according to chance with the probability of $\varepsilon$, which usually $\varepsilon$ set to a low value. Under this condition, not only the agent will act optimally but also the agent will have the chance of finding the global optimum due to explore the states by chance. Therefore, in each Q-state, the agent will choose the action accidentally with the probability of $\varepsilon$ and will choose the action according to optimal policy with probability of $1 - \varepsilon$ or follow the current policy.

## 5.3 Exploration Functions

To have a sufficient exploration, the value of each Q-state can replace with a function which models the number of observing a state and an estimation of the value of that state. For instance, an exploration function can define as follows:

$$f(u,n) = \frac{u+k}{n} \qquad (17)$$

Which in (17), $f$ is an optimistic estimation of a Q-state, $u$ is an estimation of the value of state $u$, and $n$ is the number of observing of state $u$. It is clear that each action which has better Q-state or $u$ has a higher value of $f$ as well. In other words, better actions have a higher chance of choosing. On the other hand, a higher value of $n$ will decrease the chance of choosing that action. Therefore, those states which has explored will not explore again. The normal updating rule of Q-states is achieving as follows:

$$Q(s,a) \tilde{\alpha} R(s,a,s') + \gamma \max_{a'} Q(s',a') \qquad (18)$$

According to the concepts of exploration function, the revised updating rule of Q-state will be as follows:

$$Q(s,a) \tilde{\alpha} R(s,a,s') + \gamma \max_{a'} f(Q(s',a'), N(s',a')) \qquad (19)$$

## 6 Approximate Q-Learning Algorithm

As it is mentioned in the previous section, Q-learning algorithm could be practical for the finite numbers of states and actions. It is completely obvious that in real environment, the number of states and actions are infinite, for instance, for an autonomous robot or vehicle, there are infinite numbers of possible states and actions for the agent. For these kinds of problems, using Q-learning algorithm could not a beneficial solution for determining the value of Q-states. Hence, Approximate Q-learning algorithm has proposed to handle these types of problems. In this section this algorithm has presented.

The cardinal idea of approximate Q-learning is the use of generalization rule. In other words, by observing a bounded set of some states, the agent tries to learn, and then generalize the learned things to other states. To make it clear an example of a Pacman game has provided. Consider an environment which has shown in Figure 16. According to Figure 16, if the agent is in each of the specified state, each state is considered as a different experience and the agent (Pacman) should learn each of states separately. Since the number of states are very high in this environment, solving this problem by Q-learning may take a long time due to high level of computations. By generalizing similar states to other states, these computations can decrease dramatically. For this purpose, it could be beneficial to define the environment based on the property of states in the environment. As a result, in approximate Q-learning, each state defines as vector of properties. For instance, for the example of Figure 16, the properties can define as follows:

- Distance to nearest ghost
- Distance to nearest food
- Number of ghosts
- Whether the Pacman in a Tunnel? True or False.
- And so on …

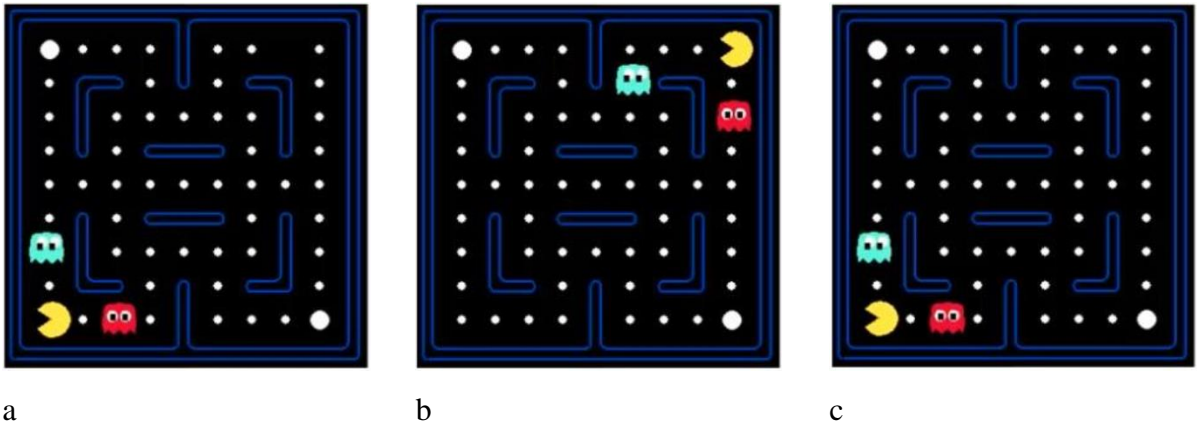

a  b  c

**Figure 16: Example of a Pacman – Concept of Generalization**

According to the aforementioned concepts, each state can define as a vector of weighted properties. This vector can determine using linear or nonlinear function. In this manuscript, linear function has considered. The value of each state can define as follows:

$$V(s) = w_1 f_1(s) + w_2 f_s(s) + \cdots + w_n f_n(s) \tag{20}$$

This property can use for calculating the value of Q-states and will be as follows:

$$Q(s,a) = w_1 f_1(s,a) + w_2 f_s(s,a) + \cdots + w_n f_n(s,a) \tag{21}$$

Therefore, the agent just need to learn the weight of each property. The procedure of learning the weights using linear functions will be as follows:

1- Setting the initial value of weights accidentally
2- Observing a new experience by a transition like $(s, a, s', r)$
3- Calculating the difference: The current estimated value of Q-state which has calculated by (20), and the estimated value of new sample is $[r + \gamma \max_{a'} Q(s', a')]$, hence, the difference will be as follows:

$$difference = \left[r + \gamma \max_{a'} Q(s', a')\right] - Q(s, a) \qquad (22)$$

4- To minimize the difference, the gradient descent can use as follows:

$$Q(s, a) \leftarrow Q(s, a) + \alpha[difference] \qquad (23)$$

Therefore, the value of weights can update according to (23), and will calculated as follows:

$$w_i \leftarrow w_i + \alpha.[difference].f_i(s, a) \qquad (24)$$

Consider the Pacman game in the grid environment which has shown in Figure 16. For this environment each Q-state is defining as follows:

$$Q(s, a) = 4f_{DOF}(s, a) - 1f_{DOG}(s, a) \qquad (25)$$

Which $f_{DOF}(s, a)$, shows the inverse distance of the Q-state to nearest food, and $f_{DOG}(s, a)$, shows the inverse distance of the Q-state to nearest ghost. In the beginning of the game, the initial weights are set accidentally, and during the process of learning, the agent will update these weights. Assume that the agent is in the state which has shown in Figure 16-a. The value of $f_{DOF}(s, a)$ and $f_{DOG}(s, a)$ by performing north action is will be:

$$f_{DOF}(s, north) = 1 \; , f_{DOG}(s, north) = 1$$

Therefore, by using (25), the value of the Q-state will be:

$$Q(s, north) = 4 - 1 = 3$$

Now the agent performs north action and according to Figure 16-a, Pacman collides with the ghost and $Q(s', north) = 0$ and gets a reward with score of -500. Therefore, the observed value of this new state and the difference value will be:

$$r + \gamma \max_{a'} Q(s', a') = -500 + 0$$

$$difference = -500 - 3 = -503$$

According to (24), the new values of weights will be:

$$w_{DOF} = 4 + \alpha[-503](1) = 2$$

$$w_{DOG} = -1 + \alpha[-503](1) = -3$$

In the above equations, for simplicity the value of learning rate has set on $\alpha = \frac{2}{503}$. Finally, the vector of Q-state with new weights will update as follows:

$$Q(s, a) = 2f_{DOF}(s, a) - 3f_{DOG}(s, a)$$

# 7 Conclusion

In this manuscript, the basic concepts of reinforcement learning have introduced. The frame work of MDP for defining an uncertain problem has presented. Some of the more common model-free algorithms of RL has investigated and the drawbacks of each of have mentioned. The idea of Q-state and Q-learning algorithm for learning dynamical environment with uncertainty without knowing the dynamic of the system has presented. It has been shown that, the approximate Q-learning can apply to dynamical system with infinite numbers of states and action.

# 8 Appendix

## 8.1 Minimization of Error in Approximate Q-learning

Minimization of error in approximate Q-learning is similar to regression problem. The vector of $Q(s, a)$ is the function that the agent want to estimate it according to the sample that it finds by exploring in the environment. For minimizing the error, we should minimize the mean square error (MSE). The MSE is defining as follows:

$$error = \frac{1}{2}\sum_i (y_i - \hat{y}_i)^2 = \frac{1}{2}\sum_i (y_i - \sum_k w_k f_k(x))^2 \tag{26}$$

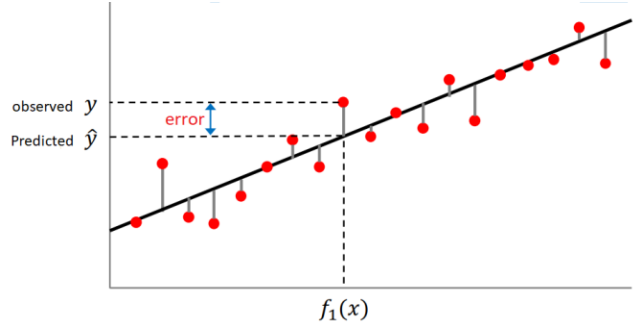

**Figure 17: Linear Regression**

In order to optimize the error which has achieved in (26), the gradient descent will use. For simplicity it has assumed that there is just a one sample, therefore, (26) will convert as follows:

$$error(w) = \frac{1}{2}(y - \sum_k w_k f_k(x))^2 \tag{27}$$

By calculating partial derivatives of (27), the following equations will achieve:

$$\frac{\partial error(w)}{\partial w_m} = -(y - \sum_k w_k f_k(x)) f_m(x)$$

$$w_m \leftarrow w_m + \alpha (y - \sum_k w_k f_k(x)) f_m(x)$$

The updating rule of weights in approximate Q-learning:

$$w_m \leftarrow w_m + \alpha \left( r + \gamma \max_{a'} Q(s', a') - Q(s, a) \right) f_m(x) \tag{28}$$